\documentclass[conference]{IEEEtran}
\IEEEoverridecommandlockouts
\usepackage{cite}
\usepackage{amsmath,amssymb,amsfonts}
\usepackage{algorithmic}
\usepackage{graphicx}
\usepackage{textcomp}
\usepackage{xcolor}
\usepackage{caption}
\usepackage[skins]{tcolorbox} 
\usepackage{multicol}
\usepackage{siunitx, makecell}
\usepackage{mathtools}
\usepackage{graphics} 
\usepackage{epsfig} 
\usepackage{mathptmx} 
\usepackage{times} 
\usepackage{amsmath} 
\usepackage{amssymb}  
\usepackage{graphicx}
\usepackage{epstopdf}
\usepackage{pdfpages}
\usepackage{cite}
\usepackage{subfigure}
\usepackage{amsbsy}
\usepackage{bm}
\usepackage{url}
\usepackage{accents}

\makeatletter 
\newcommand{\linebreakand}{%
  \end{@IEEEauthorhalign}
  \hfill\mbox{}\par
  \mbox{}\hfill\begin{@IEEEauthorhalign}
}
\makeatother 

\begin{document}

\title{
Comparative Evaluation of RGB-D SLAM Methods for Humanoid Robot Localization and Mapping
}

\author{\IEEEauthorblockN{Amirhosein Vedadi}
\IEEEauthorblockA{\textit{Center of Advanced Systems} \\ \textit{and Technologies (CAST)} \\
\textit{School of Mechanical Engineering} \\
\textit{University of Tehran}\\
amirhosein.vedadi@ut.ac.ir}
\and
\IEEEauthorblockN{Aghil Yousefi-Koma}
\IEEEauthorblockA{\textit{Center of Advanced Systems} \\ \textit{and Technologies (CAST)} \\
\textit{School of Mechanical Engineering} \\
\textit{University of Tehran}\\
aykoma@ut.ac.ir}
\and
\IEEEauthorblockN{Parsa Yazdankhah}
\IEEEauthorblockA{\textit{Center of Advanced Systems} \\ \textit{and Technologies (CAST)} \\
\textit{School of Mechanical Engineering} \\
\textit{University of Tehran}\\
parsa.yazdankhah@ut.ac.ir}
\and

\linebreakand 

\IEEEauthorblockN{Amin Mozayyan}
\IEEEauthorblockA{\textit{Center of Advanced Systems} \\ \textit{and Technologies (CAST)} \\
\textit{School of Mechanical Engineering} \\
\textit{University of Tehran}\\
aminmozayyan@ut.ac.ir}
}

\maketitle

\begin{abstract}
In this paper, we conducted a comparative evaluation of three RGB-D SLAM (Simultaneous Localization and Mapping) algorithms—RTAB-Map, ORB-SLAM3, and OpenVSLAM—for SURENA-V humanoid robot localization and mapping.
Our test involves the robot to follow a full circular pattern, with an Intel® RealSense™ D435 RGB-D camera installed on its head.
In assessing localization accuracy, ORB-SLAM3 outperformed the others with an ATE of 0.1073, followed by RTAB-Map at 0.1641 and OpenVSLAM at 0.1847.
However, it should be noted that both ORB-SLAM3 and OpenVSLAM faced challenges in maintaining accurate odometry when the robot encountered a wall with limited feature points. Nevertheless, OpenVSLAM demonstrated the ability to detect loop closures and successfully relocalize itself within the map when the robot approached its initial location.
The investigation also extended to mapping capabilities, where RTAB-Map excelled by offering diverse mapping outputs, including dense, OctoMap, and occupancy grid maps. In contrast, both ORB-SLAM3 and OpenVSLAM provided only sparse maps.

\end{abstract}

\begin{IEEEkeywords}
\textit{RGB-D SLAM, humanoid robot, localization, mapping, loop closure detection}
\end{IEEEkeywords}

\section{Introduction}
With recent advancements in the development of more reliable and stable bipedal platforms, there is a growing need to enable these robots to autonomously operate without human intervention. This necessitates the capability of these robots to perceive and understand their surrounding environment.

The field of simultaneous localization and mapping (SLAM) has witnessed significant advancements in recent years. SLAM algorithms play a crucial role in enabling robots to perceive their environment and autonomously navigate within it. These algorithms allow robots to simultaneously build a map of their surroundings while estimating their own pose or location within that map.

\begin{figure}[tbp]
 \centering
 \subfigure[]{\includegraphics[width=0.24\textwidth,height=5.7cm]{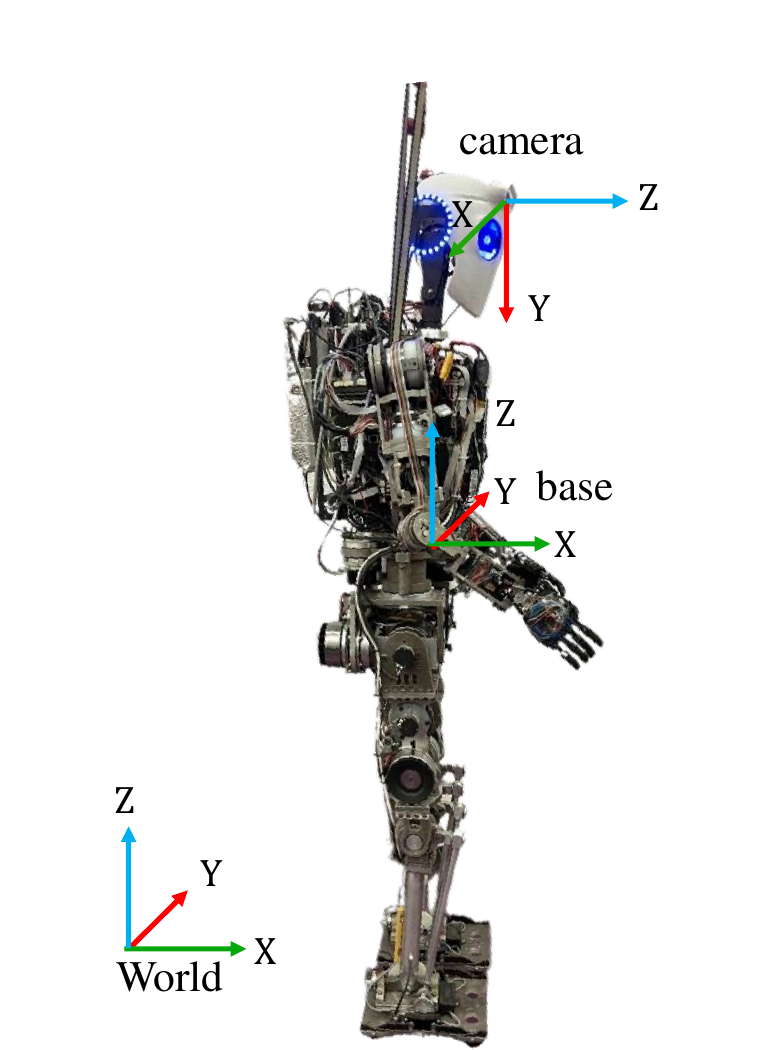}}
 \subfigure[]{\includegraphics[width=0.24\textwidth,height=5cm]{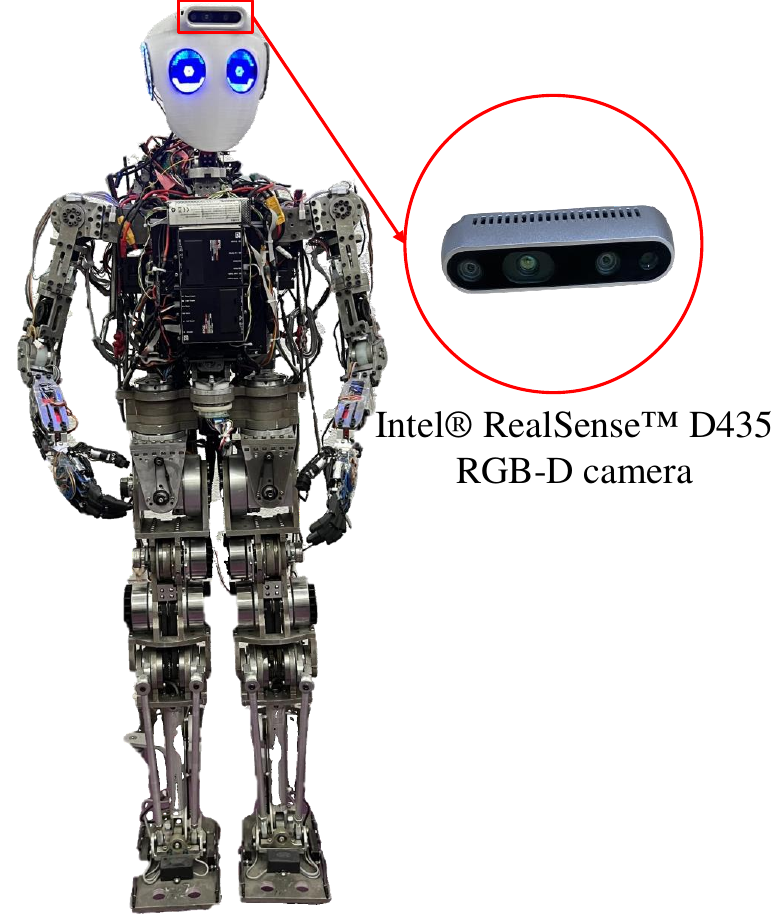}}
 \caption{(a) World, base, and camera co-ordinate frames of SLAM algorithms (b) Intel® RealSense™ D435 RGB-D camera placement on the SURENA-V humanoid robot}
 \label{fig1}
\end{figure}

Mapping the environment and simultaneously determining the robot's position can be achieved using various sensors, including visual ones like cameras or laser sensors such as LiDAR \cite{Cadena2016SLAMfuture}. Visual-based SLAM algorithms can be considered especially attractive, due to their sensor configuration simplicity, miniaturized size, and low cost \cite{Barros2022ACS}. The visual-based SLAM approaches can be divided into three main categories: monocular, stereo, and RGB-D. These approaches are then further divided into feature-based (matching visual features over a number of frames) and optical flow techniques (based on the intensity of all pixels or specific regions in sequential images) \cite{Yousif2015AnOT}.

"ORB-SLAM" is a method based on pose-graph optimization with an integrated loop closure detection algorithm \cite{mur2015orb}. This method was later presented for depth and stereo cameras, which are of interest in this research \cite{mur2017orb}. Furthermore, 
recent developments have led to the introduction of the third generation of this method, named "ORB-SLAM3", which incorporates visual-inertial SLAM capabilities and supports multiple maps simultaneously.\cite{campos2021orb}. 

The "RTAB-Map" algorithm is another SLAM algorithm, distinguishing itself by its comprehensive approach that fully addresses essential SLAM components in addition to memory management policies, which leads to optimized performance \cite{labbe2019rtab}. This algorithm is adaptable to different combinations of visual and laser based sensors.
Having an external odometry node, which enables the use of additional ranging techniques to enhance the performance of the SLAM algorithm, is one of the key advantages of this approach. Moreover, the RTAB-Map algorithm employs a visual odometry technique if the robot is not equipped with any specific odometry techniques \cite{scaramuzza2011visual}.

SLAM algorithms are essential in humanoid robotics field, empowering robots to map their surroundings and navigate using sensors like cameras and lasers. This capability enhances their autonomy and adaptability, making them valuable assets in diverse applications.

Nevertheless, implementing a visual SLAM on a humanoid robot has challenging problems, including inherent factors such as erratic swaying motion and uncertainties\cite{ahn2012board}. Furthermore, the reliability of odometry is compromised by feet slipping\cite{wirbel2013humanoid}. The presence of numerous degrees of freedom in these robots amplifies the complexity of estimating their pose, which is critical for measuring angles and distances in the surrounding environment\cite{gouda2013vision}. Sharp accelerations during walking induce motion blur in images captured by the robot's camera. Additionally, the robot might just point the camera at visually featureless blank walls. This doesn't provide a meaningful framework for localization \cite{scona2017direct}.

One of the earliest attempts to implement SLAM algorithms in humanoid robots was conducted by Stasse et al. \cite{stasse2006real}. They applied their technique to the "HRP2" robot using the robot's IMU sensors and trajectory planner data. Scona et al. implemented the SLAM algorithm on the "Valkyrie" robot \cite{scona2017direct}. For a stereo camera, they utilized the direct method rather than feature-based approaches, in addition to IMU sensor data, which was also used in their estimation. Hourdakis et al. \cite{hourdakis2021roboslam} implemented the SLAM algorithm with a depth camera on the humanoid robot "NAO" and were able to create a detailed map of the surrounding area.

Zhang et al. \cite{zhang2018dense} implemented the "ElasticFusion" \cite{whelan2015elasticfusion} method on the humanoid robot "HRP4" in an environment assuming the presence of humans. In this research, humans, who are considered as dynamic objects in the scene, are identified with the help of deep learning methods and removed from the map.

Leveraging feedback from SLAM algorithm outputs for robot control has been explored in the research conducted by Tanguy et al. \cite{tanguy2019closed}. In their study, they integrated the algorithm's output with the robot's force-torque sensors within a predictive model control framework. Their efforts, carried out on the "HRP4" robot, resulted in notable improvements in its movement under external stresses and disturbances.

In this paper, we present a rigorous evaluation of various RGB-D SLAM methods, considering their accuracy, efficiency, robustness, scalability, and adaptability to humanoid walking scenarios. We analyze prominent techniques such as RTAB-Map, ORB-SLAM3, and OpenVSLAM, providing a comprehensive overview of their underlying principles and algorithmic frameworks. Additionally, the experimental setups employed for evaluation are being discussed, ensuring the reliability and reproducibility of the comparative analysis.

In Section II, we conduct a review of the employed SLAM methods and provide detailed information about each of them. In Section III, we present the experimental evaluations conducted on the SuURENA-V humanoid robot, comparing the results obtained from different algorithms. Finally, in Section IV, we provide a conclusive summary of the paper.

\renewcommand{\arraystretch}{1.3}
\setlength{\arrayrulewidth}{0.2mm}
\setlength{\tabcolsep}{18pt}
\section{Visual Slam Methods}

    \begin{table*}[t]
        \begin{center}
            \caption{Comparison of representative visual slam methods}
            \label{tab2}
            \begin{tabular}{|| c | c | c | c ||}
                \cline{2-4}
                \multicolumn{1}{c|}{} & RTAB-Map & ORB-SLAM3 & OpenVSLAM \\
                \hline
                ROS Compatible & \checkmark & \checkmark & \checkmark \\
                Feature Matching & GFTT, BRIEF & FAST, BRIEF & FAST, BRIEF \\
                Map Output &  \thead{Dense, OctoMap, \\Occupancy Grid} & Sparse & Sparse   \\ 
                Loop Closure & \checkmark & \checkmark & \checkmark   \\ 
                Camera Tracking & Indirect & Indirect & Indirect   \\ 
                Implementation & C++ & C++ & C++  \\ 
                Publication Year & 2019 & 2021 & 2019   \\ 
                Sensor Type & \thead{Mono, Stereo, RGB-D, Fisheye,\\LiDAR, External Odometry} & \thead{Mono, Stereo, RGB-D,\\Fisheye, IMU} & \thead{Mono, Stereo, RGB-D,\\Fisheye, Equirectangular}  \\ 
                \hline
            \end{tabular}
        \end{center}
    \end{table*}

In the following section, the three mentioned SLAM methods will be introduced which are briefly compared in Table I.
\subsection{RTAB-Map}
RTAB-Map is a graph-based SLAM algorithm. One of its prominent strengths lies in its compatibility with the Robot Operating System (ROS), making it a valuable choice for robotics applications. This algorithm exhibits versatility by accommodating diverse input sensors, including RGB-D and stereo cameras, as well as lidar. Furthermore, it offers the flexibility to incorporate external odometry sources, such as encoder odometry or global positioning systems (GPS), enabling fusion of multi-modal sensor data.

Given the potential discrepancy in data publishing frequencies across these sensors, RTAB-Map incorporates a synchronization module to ensure proper alignment and temporal consistency. Moreover, RTAB-Map includes a dedicated component for efficient memory management, which assumes paramount importance in scenarios involving large-scale environments. This memory management module effectively handles the growth of the graph structure and optimizes computation time, contributing to the algorithm's scalability and robustness.

In the absence of external odometry information, RTAB-Map employs visual odometry as an alternative means of estimating camera motion. By default, RTAB-Map utilizes the Good Features to Track (GFTT) algorithm to identify distinctive visual features that are suitable for tracking purposes. To describe and match these tracked features, RTAB-Map employs the Binary Robust Independent Elementary Features (BRIEF) descriptor. In order to reduce the search space and improve computational efficiency during feature matching, RTAB-Map employs a constant velocity motion model. Once the features are matched between frames, RTAB-Map utilizes the Perspective-n-Point (PnP) algorithm to estimate the relative motion of the camera with respect to the previous keyframe.

RTAB-Map employs a bag-of-words method for loop closure detection, a critical component in SLAM algorithms. The bag-of-words approach involves creating a visual vocabulary or dictionary that represents different visual patterns or words. These visual words are generated by clustering the features extracted from the visual odometry process. During loop closure detection, the algorithm compares the features of the current frame with the visual words stored in the map. By matching the features of the current frame with those in the map, potential loop closures or revisits to previously visited locations can be identified.

After detecting a loop closure, RTAB-Map applies the g2o graph optimization to minimize the error between the map and odometry. However, it is important to note that if the resulting error after optimization remains high, the loop closure may not be considered valid or reliable. The g2o graph optimization aims to refine the estimated poses and landmarks in the map by iteratively minimizing the error residuals. This optimization process seeks to align the loop closure constraints with the rest of the odometry information to achieve a more accurate and consistent map representation.

During the mapping phase, RTAB-Map has the capability to generate different types of maps, including dense maps, occupancy maps, and sparse maps, based on the specific requirements and objectives of the application.
\subsection{ORB-SLAM3}
ORB-SLAM3 is an openly available software library designed for SLAM tasks, compatible with a variety of camera types and lens models, such as monocular, stereo, and RGB-D cameras, that is released in 2020 \cite{campos2021orb}. It stands out as the first system capable of performing visual-inertial SLAM using a maximum-a-posteriori estimation approach. Within ORB-SLAM3, there exists a multi-map representation referred to as the "Atlas," which consists of two types of maps: active and nonactive. The active map is primarily utilized for the incoming frame localization, while the nonactive maps have various purposes, including relocalization, loop closure, and map merging.

\noindent The system functions through three main threads:
\begin{enumerate}
\item Tracking thread is responsible for processing sensor data, determining the pose of the current frame, and minimizing errors in projecting matched map features. In the visual-inertial mode, it additionally estimates parameters like body velocity and IMU biases and can relocalize frames within all available maps when tracking is disrupted.
\item Local mapping thread handles tasks such as incorporating keyframes and points into the active map, eliminating redundant data, refining the map, and initializing IMU parameters.
\item Loop and map merging thread is tasked with identifying common regions between the active map and the entire Atlas as keyframes are generated. If the shared area is part of the active map, it corrects any loops. If the common area belongs to a different map, it seamlessly merges both maps into a single entity, which then becomes the new active map.
\end{enumerate}

\subsection{OpenVSLAM}
OpenVSLAM, a modular and versatile open-source library, is easy to use and extend by users of visual SLAM. This software framework, released in 2019, is specifically tailored for visual SLAM and visual odometry tasks
\cite{sumikura2019openvslam}.
It exhibits the ability to handle various camera models, including perspective, fisheye, and equirectangular, and can be customized to accommodate optional camera models. It is compatible with monocular, stereo, and RGB-D camera setups, and it offers the capability to store and load created maps, which can subsequently be utilized for image localization using prebuilt maps. It is a cross-platform viewer that can be used on multiple platforms and is accessible through web browsers.

 Inspired by the Structure from Motion (SfM) frameworks, OpenVSLAM is provided, which is compatible with various types of camera models. Thus, the aim is to enhance the usability and extensibility of visual SLAM for 3D mapping and localization.

\begin{figure*}[tbp]
 \centering
 \subfigure[RTAB-Map]{\includegraphics[trim={2cm 8cm 2cm 9cm},clip,width=0.3\textwidth]{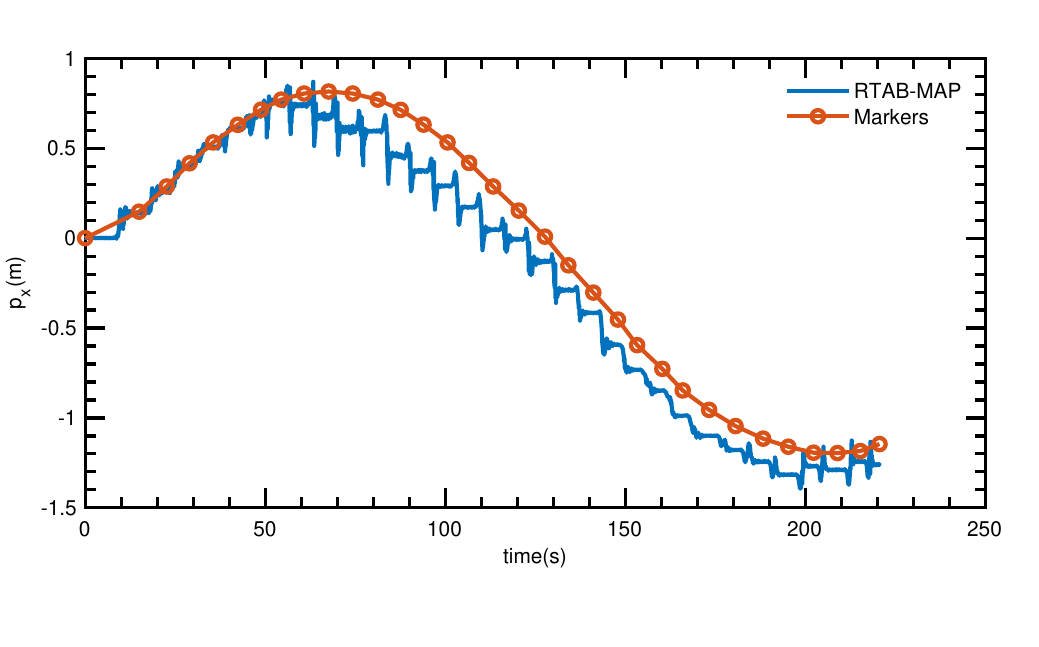}}
 \subfigure[OpenVSLAM]{\includegraphics[trim={2cm 8cm 2cm 9cm},clip,width=0.3\textwidth]{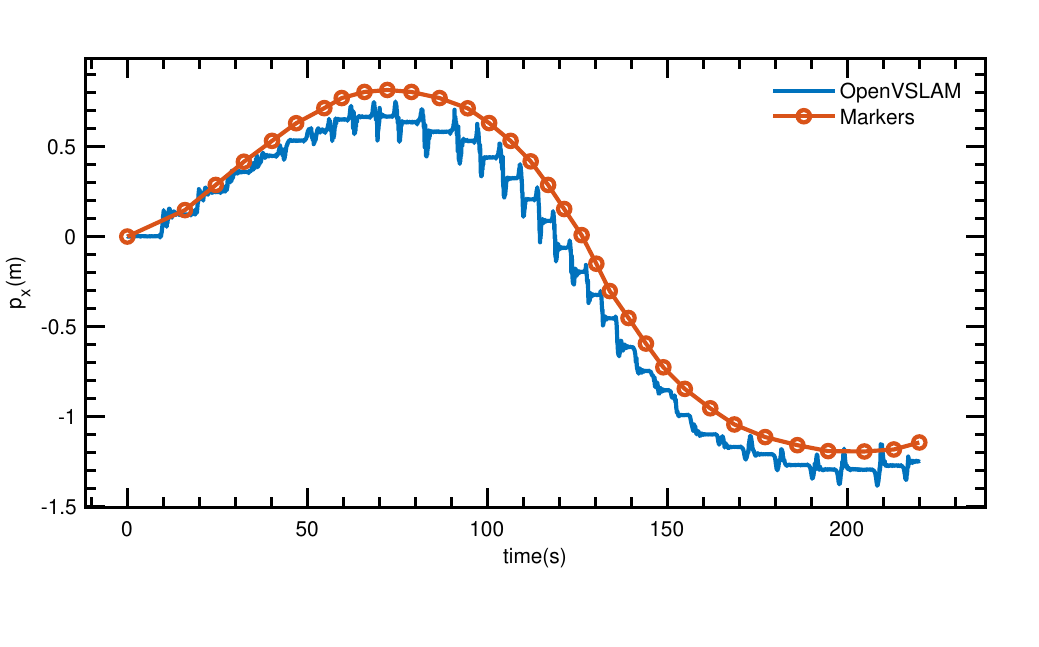}}
 \subfigure[ORB-SLAM3]{\includegraphics[trim={2cm 8cm 2cm 9cm},clip,width=0.3\textwidth]{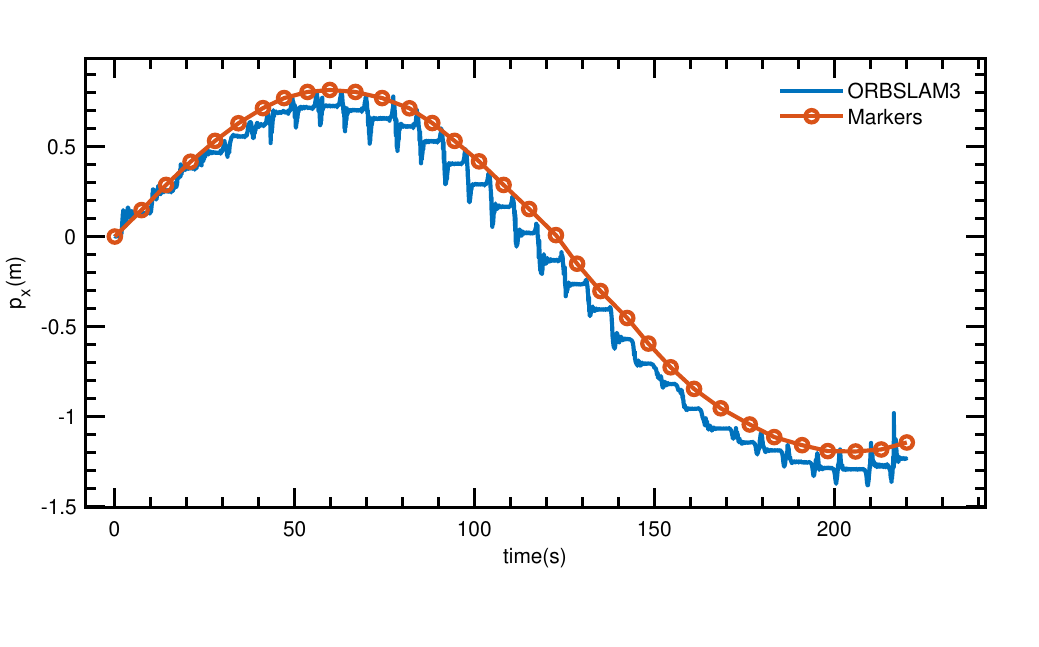}}
 \subfigure[RTAB-Map]{\includegraphics[trim={2cm 8cm 2cm 9cm},clip,width=0.3\textwidth]{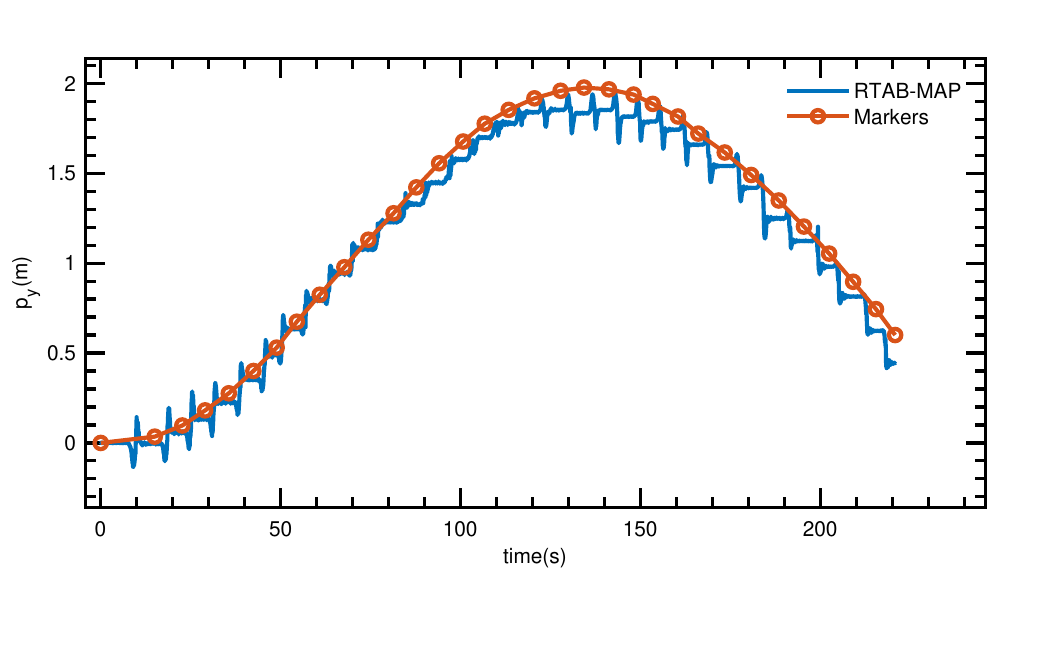}}
 \subfigure[OpenVSLAM]{\includegraphics[trim={2cm 8cm 2cm 9cm},clip,width=0.3\textwidth]{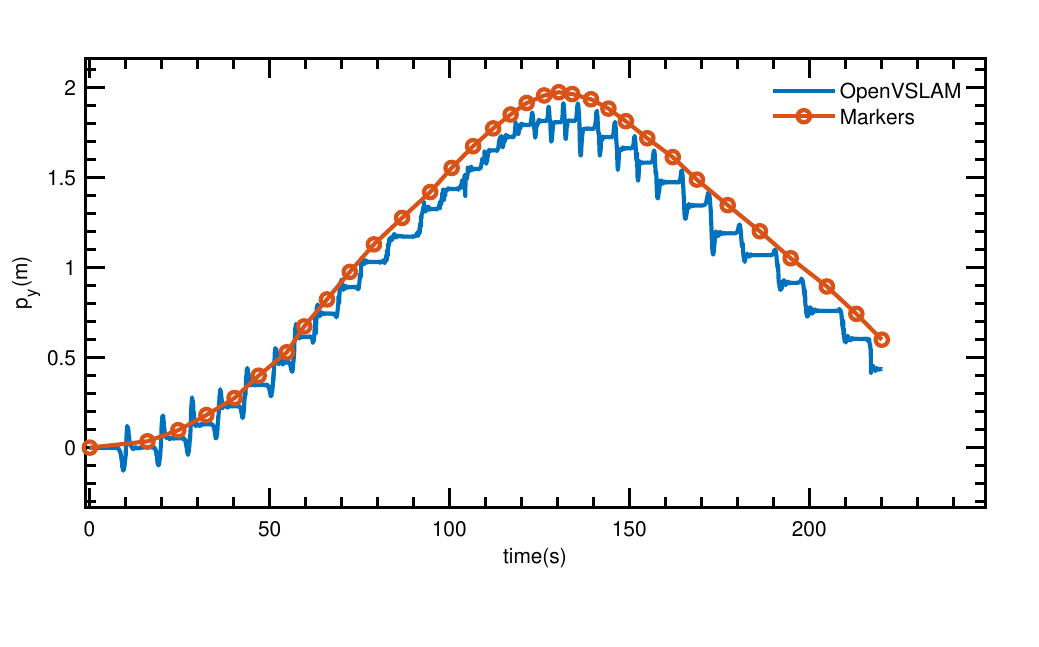}}
 \subfigure[ORB-SLAM3]{\includegraphics[trim={2cm 8cm 2cm 9cm},clip,width=0.3\textwidth]{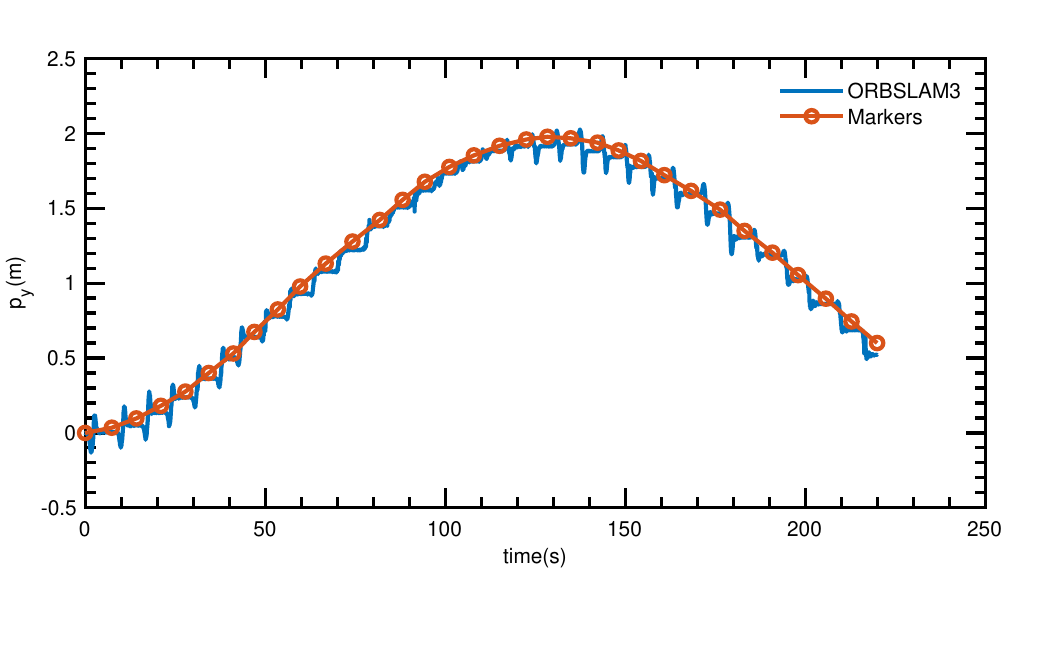}}

 \caption{Localization outputs of algorithms in comparison with ground truth marker positions in x and y directions}
 \label{fig5}
\end{figure*}

OpenVSLAM utilizes a graph-based SLAM algorithm, employing the indirect method and ORB \cite{rublee2011orb} as the selected feature extractor. The library is organized into three modules: tracking, mapping, and global optimization. The tracking module is responsible for estimating the camera pose for each sequentially inputted frame, achieved through keypoint matching and pose optimization. It also makes decisions about inserting new keyframes. The mapping module is in charge of creating and expanding the map with new 3D points, along with performing windowed map optimization, referred to as local bundle adjustment. The global optimization module handles loop detection, pose-graph optimization, and global bundle adjustment.
Pose-graph optimization, implemented with g2o \cite{kummerle2011g}, is employed to address and resolve trajectory drift.
In terms of capabilities, OpenVSLAM is comparable to ORB-SLAM, particularly in its ability to perform SLAM using an equirectangular camera. This feature proves highly advantageous for tracking and mapping, as it provides an omnidirectional view unlike the limited perspective view offered by ORB-SLAM. Furthermore, OpenVSLAM allows for the storage and loading of created maps, a functionality not present in ORB-SLAM. Lastly, OpenVSLAM boasts a shorter tracking time compared to ORB-SLAM, primarily due to the more optimized implementation of ORB extraction within OpenVSLAM.

\begin{figure}[htbp]
 \centering
 \includegraphics[trim={2cm 9cm 2cm 9cm},clip,width=0.47\textwidth, ]{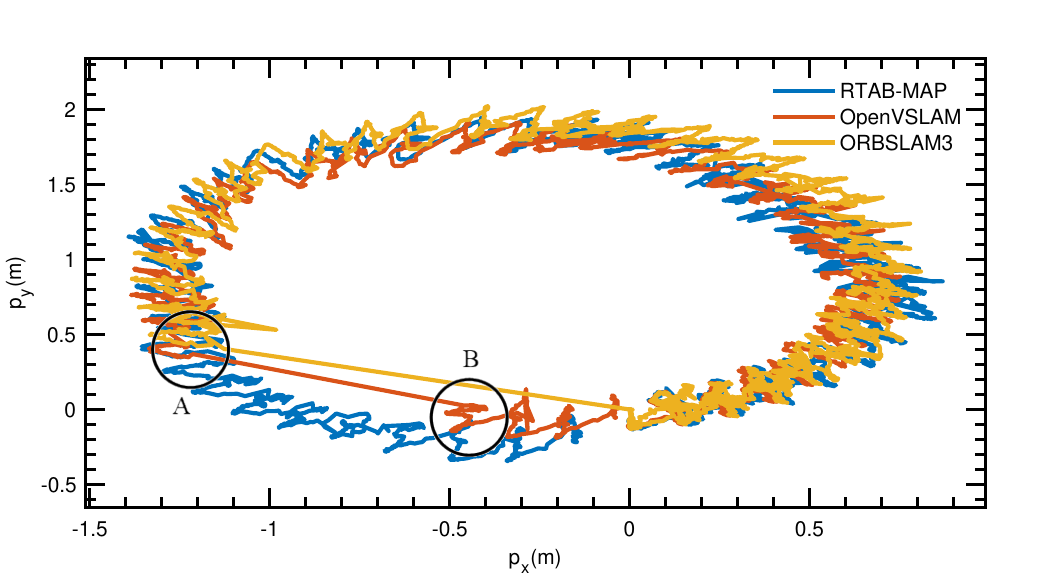}
 \caption{Comparison of odometry output of algorithms. As shown, RTAB-Map has successfully tracked the robot's position; however, the other two methods lost track of the robot just after the number of features were diminished in the environment (A). OpenVSLAM, on the other hand was able to relocalize itself in the map after detecting a loop closure (B).}
 \label{fig3}
\end{figure}

\section{Experiments And Results}

\subsection{System Overview}

\begin{figure*}[htbp]
 \centering
 \subfigure[]{\includegraphics[height=0.15\textheight]{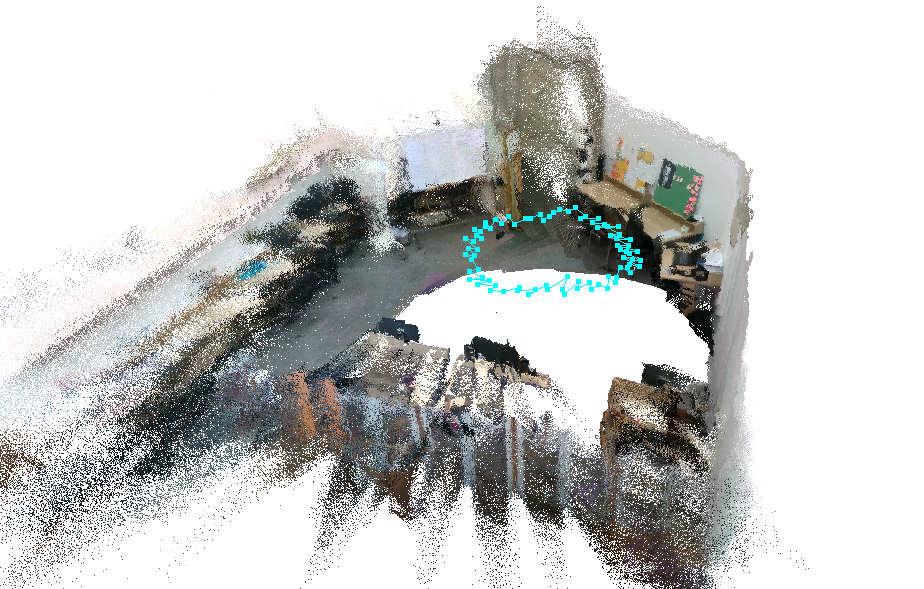}}
 \subfigure[]{\includegraphics[height=0.15\textheight]{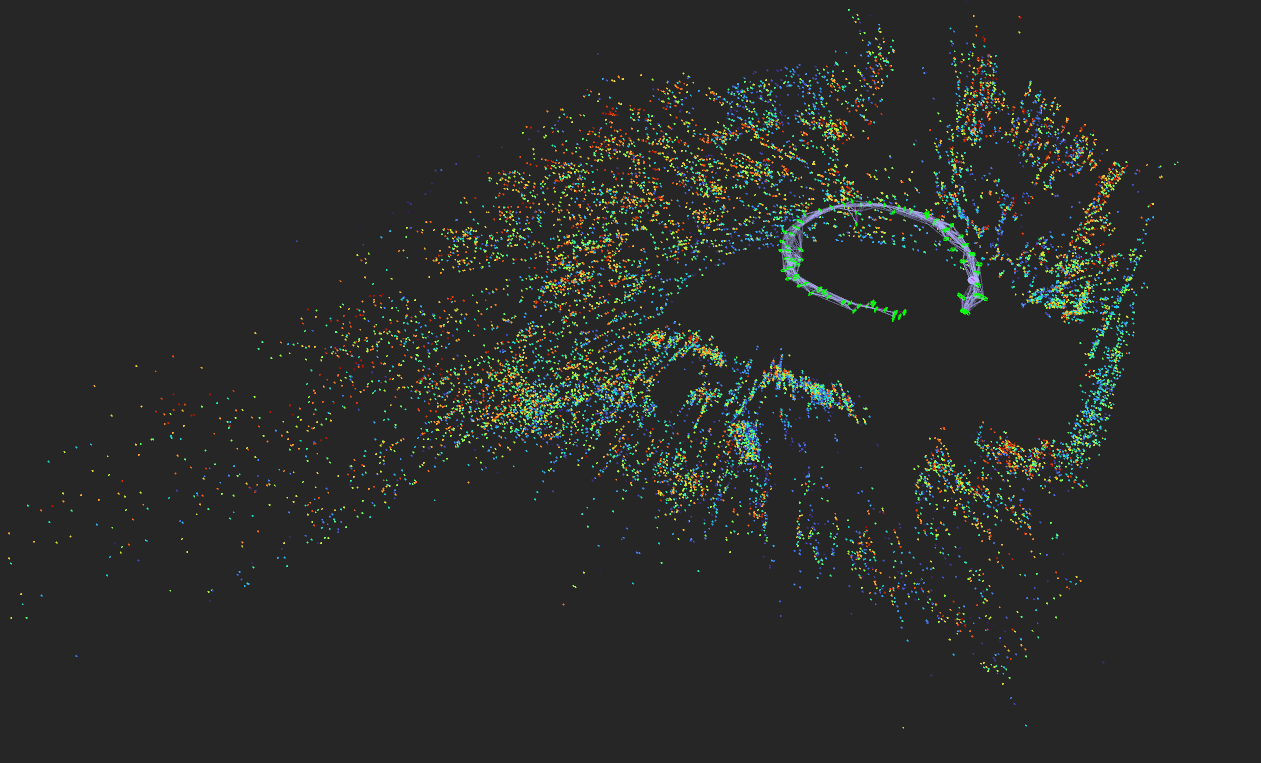}}
 \subfigure[]{\includegraphics[height=0.15\textheight]{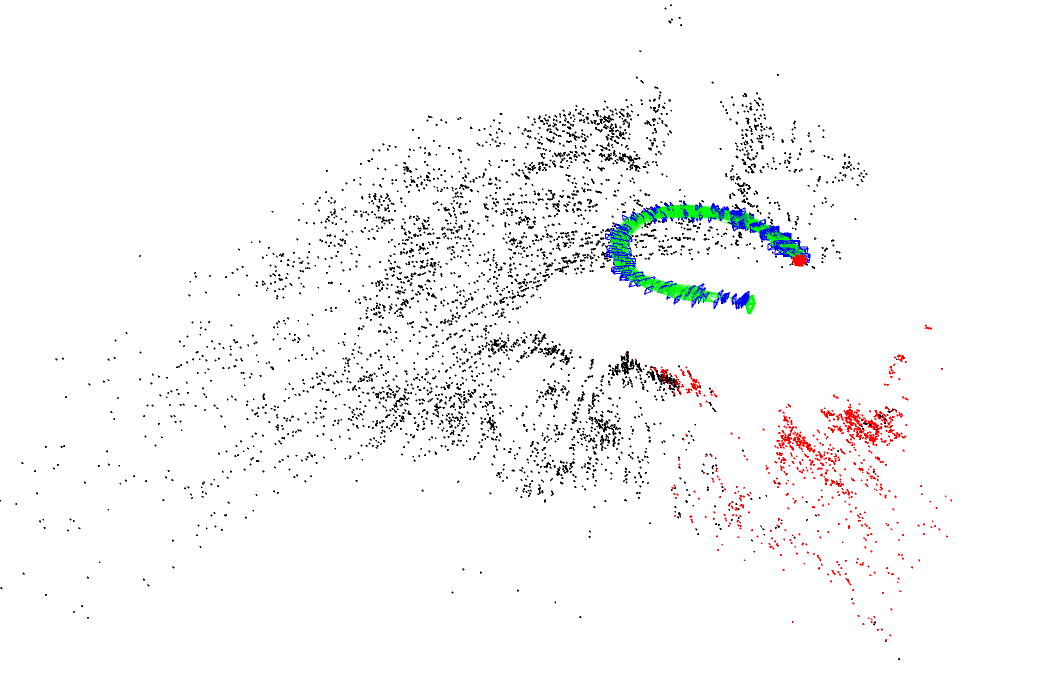}}
 \caption{Output maps of investigated methods (a) Dense map of RTAB-Map algorithm (b) Sparse map of OpenVSLAM algorithm (c) Sparse map of ORB-SLAM3 algorithm}
 \label{fig4}
\end{figure*}

\begin{figure*}[tbp]
 \centering
 \subfigure[]{\includegraphics[width=0.32\textwidth]{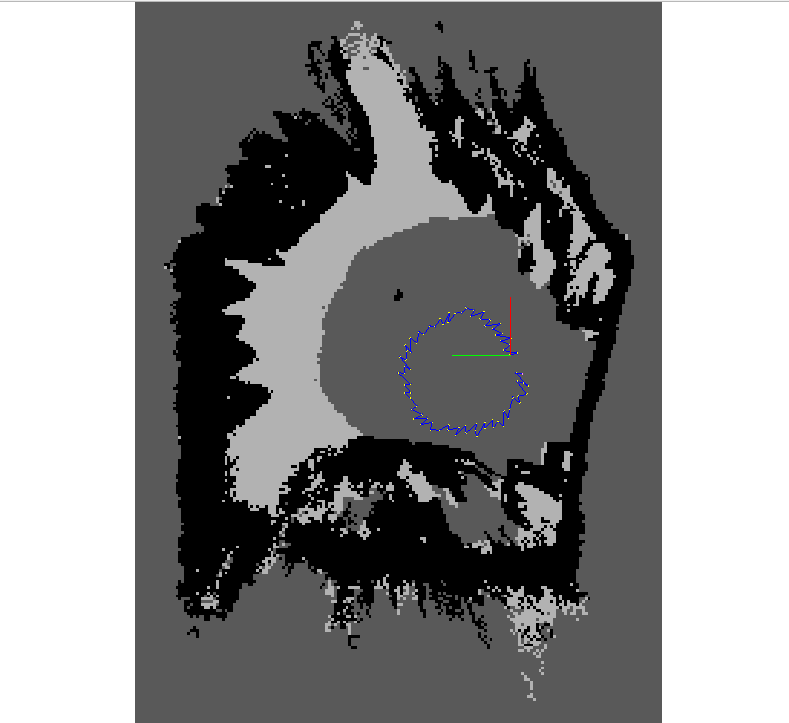}}
 \subfigure[]{\includegraphics[width=0.3\textwidth]{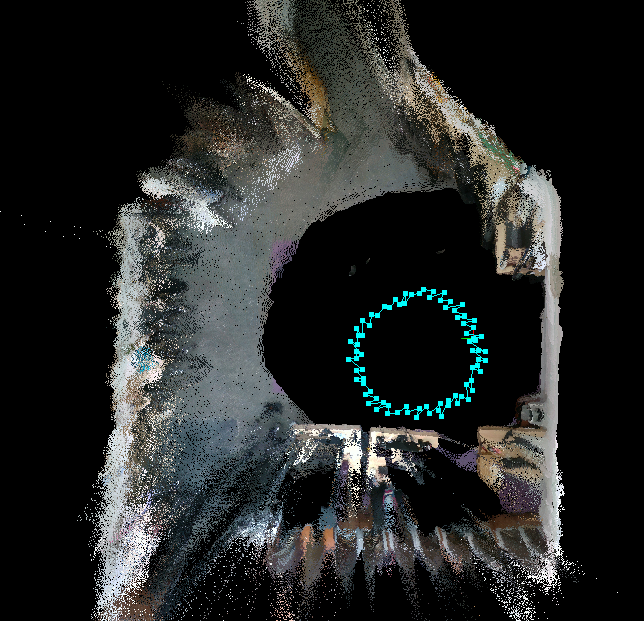}}
 \caption{RTAB-Map outputs (a) occupancy grid (b) bird's eye view Dense map}
 \label{fig2}
\end{figure*}

The experiment utilized the SURENA-V humanoid robot, developed by the Center of Advanced Systems and Technologies (CAST), as the hardware platform. This robot possesses 44 degrees of freedom. The robot stands at a height of 168 cm and weighs 68 kg. Additionally, the robot is equipped with an Intel® RealSense™ D435 RGB-D camera, which is installed on its head as shown in Fig. 1(b). The camera lenses have a field of view measuring 87° × 58°.
The camera is connected to a laptop with a 4-core Intel Core i7-11370H CPU, 8 GB of GDDR6 system memory, and an Nvidia GeForce RTX 3070 GPU.
A keyboard interface is employed to control the SURENA humanoid robot as it walks on a flat floor, with a control frequency of 200 Hz. The robot follows a circular path, rotating by 0.17 radians and moving 0.15 meters per step. The robot trajectory was obtained from optimized parameters based on \cite{9663522}, and the control structure of \cite{10000072} was used to reduce the robot's feet ground impact during walking.
The dataset was generated by utilizing an RGB-D camera, which consists of RGB and depth images captured at a resolution of 640 × 480 pixels and a frame rate of 30 fps.

\subsection{Evaluation of Localization}
For the purpose of evaluating localization accuracy, we utilized a marker-based method to establish a reliable ground truth reference.
Following each step taken by the robot and upon reaching a stable state, we placed markers on the ground. Subsequently, we measured the positions and orientations of these markers after the experiment to determine the robot's location and orientation at the end of each step. Finally, we compared the algorithm outputs at these marker locations with the corresponding measured positions.
 Additionally, we calculated the Absolute Trajectory Error (ATE) at these points for each algorithm using the following equation :
\begin{equation}
	ATE = \sqrt{\frac{1}{N} \sum_{i=1}^{N} \left \| log(T_{gt, i}^{-1}T_{es, i})^\vee \right \|^2}
\end{equation}
where $N$ represents the total number of timestamps, $T_{gt, i}$ denotes the ground truth rigid body transform at the time i, and $T_{es, i}$ represents the estimated rigid body transform at the time i. Also, the $()^\vee$ operator is a mathematical operator used to transform a skew-symmetric error matrix into a vector space representation. 

\begin{table}[]
    \begin{center}
        \caption{Comparison of Absolute Trajectory Error (ATE) between VSLAM algorithms based on the End of Step position}
        \resizebox{0.3\textwidth}{!}{$
        \begin{tabular}{l|c||}
\cline{2-2}
                                & $ATE_{eos}(m)$ \\ \hline
\multicolumn{1}{||l|}{RTAB-MAP}  & 0.1641        \\
\multicolumn{1}{||l|}{ORBSLAM3}  & 0.1073        \\
\multicolumn{1}{||l|}{OpenVSLAM} & 0.1847        \\ \hline
\end{tabular}
        $}
    \end{center}
\end{table}
Table II presents the results of ATE computation, indicating the localization accuracy of the evaluated methods. As observed, ORBSLAM3 demonstrates superior performance in terms of localization accuracy, followed by RTAB-Map and OpenVSLAM.
\subsection{Evaluation in Challenging Settings and Loop Closure}
One of the primary challenges encountered when implementing SLAM on humanoid robots is the inherent movement and shaking of the robot during walking. However, all three algorithms demonstrate satisfactory performance in localization under these conditions, provided they have access to an adequate number of features.

Another notable challenge faced by feature-based SLAM algorithms arises in environments characterized by a scarcity of distinct visual features. During a specific segment of our experiment, the robot encountered a white wall with limited features. In this particular scenario, only the RTAB-Map algorithm managed to maintain odometry, while the other two algorithms experienced difficulties and lost their position. This situation is visually depicted in Fig. 3.

Once the robot returned to a position approximately close to its initial starting point, both the RTAB-Map and OpenVSLAM algorithms successfully detected loop closures. Following the detection of a loop closure, the OpenVSLAM algorithm demonstrates its ability to relocalize itself within the previously constructed map.

\subsection{Evaluation of Mapping}
During the mapping phase, both ORB-SLAM3 and OpenVSLAM exhibit the ability to generate sparse feature maps. The resulting maps, depicted in Figure 4, were obtained prior to the loss of odometry by the algorithms. These maps provide a visual representation of the environment captured by the robot, highlighting the distribution and density of the detected features in the scene. 

RTAB-Map, unlike ORB-SLAM3 and OpenVSLAM, offers a wider range of options in the mapping phase. As depicted in Figure 4, RTAB-Map is capable of generating a dense map of the robot's environment in real-time. This dense map provides a more detailed representation of the surroundings and captures a higher density of features.

Additionally, in Fig. 5, we present the occupancy grid map generated by RTAB-Map. This type of map is particularly useful for robot navigation tasks, as it provides a binary representation of the environment, indicating the presence or absence of obstacles. The occupancy grid map can aid in path planning, obstacle avoidance, and other navigation-related tasks, enhancing the robot's autonomous capabilities.

\section{Conclusion}

The paper explored three prominent RGB-D SLAM algorithms: RTAB-Map, ORB-SLAM3, and OpenVSLAM. 
Firstly, the algorithms showcased their capabilities in localization accuracy. The evaluation revealed that ORB-SLAM3 demonstrated superior performance in terms of Absolute Trajectory Error (ATE) during robot movement, followed by RTAB-Map and OpenVSLAM, respectively.
Furthermore, loop closure detection was examined. RTAB-Map has successfully tracked the robot's position and maintained it on the track even when the robot encountered a wall upon itself and the number of features decreased drastically. The other two methods lost track of the robot in this situation; however, the OpenVSLAM had partially refined this problem by relocalizing itself in the map after detecting a loop closure. On the other hand, ORB-SLAM3 was unable to detect loop closure and did not follow the robot's trajectory after losing track in the first place.
Mapping capabilities were also investigated. RTAB-Map represented impressive performance, revealing diverse outputs such as dense, OctoMap and occupancy grid maps. The other two methods were only provided with sparse maps.

\bibliographystyle{IEEEtran}
\bibliography{bibliography}

\begin{thebibliography}{10}
\providecommand{\url}[1]{#1}
\csname url@samestyle\endcsname
\providecommand{\newblock}{\relax}
\providecommand{\bibinfo}[2]{#2}
\providecommand{\BIBentrySTDinterwordspacing}{\spaceskip=0pt\relax}
\providecommand{\BIBentryALTinterwordstretchfactor}{4}
\providecommand{\BIBentryALTinterwordspacing}{\spaceskip=\fontdimen2\font plus
\BIBentryALTinterwordstretchfactor\fontdimen3\font minus \fontdimen4\font\relax}
\providecommand{\BIBforeignlanguage}[2]{{%
\expandafter\ifx\csname l@#1\endcsname\relax
\typeout{** WARNING: IEEEtran.bst: No hyphenation pattern has been}%
\typeout{** loaded for the language `#1'. Using the pattern for}%
\typeout{** the default language instead.}%
\else
\language=\csname l@#1\endcsname
\fi
#2}}
\providecommand{\BIBdecl}{\relax}
\BIBdecl

\bibitem{Cadena2016SLAMfuture}
C.~Cadena, L.~Carlone, H.~Carrillo, Y.~Latif, D.~Scaramuzza, J.~Neira, I.~Reid, and J.~Leonard, ``Past, present, and future of simultaneous localization and mapping: Towards the robust-perception age,'' \emph{{IEEE Transactions on Robotics}}, vol.~32, no.~6, p. 1309–1332, 2016.

\bibitem{Barros2022ACS}
A.~M. Barros, M.~Michel, Y.~Moline, G.~Corre, and F.~Carrel, ``A comprehensive survey of visual slam algorithms,'' \emph{Robotics}.

\bibitem{Yousif2015AnOT}
K.~Yousif, A.~Bab-Hadiashar, and R.~Hoseinnezhad, ``An overview to visual odometry and visual slam: Applications to mobile robotics,'' \emph{Intelligent Industrial Systems}, vol.~1, pp. 289--311, 2015.

\bibitem{mur2015orb}
R.~Mur-Artal, J.~M.~M. Montiel, and J.~D. Tardos, ``Orb-slam: a versatile and accurate monocular slam system,'' \emph{IEEE transactions on robotics}, 2015.

\bibitem{mur2017orb}
R.~Mur-Artal and J.~D. Tard{\'o}s, ``Orb-slam2: An open-source slam system for monocular, stereo, and rgb-d cameras,'' \emph{IEEE transactions on robotics}, vol.~33, no.~5, pp. 1255--1262, 2017.

\bibitem{campos2021orb}
C.~Campos, R.~Elvira, J.~J.~G. Rodr{\'\i}guez, J.~M. Montiel, and J.~D. Tard{\'o}s, ``Orb-slam3: An accurate open-source library for visual, visual--inertial, and multimap slam,'' \emph{IEEE Transactions on Robotics}, 2021.

\bibitem{labbe2019rtab}
M.~Labb{\'e} and F.~Michaud, ``Rtab-map as an open-source lidar and visual simultaneous localization and mapping library for large-scale and long-term online operation,'' \emph{Journal of Field Robotics}, 2019.

\bibitem{scaramuzza2011visual}
D.~Scaramuzza and F.~Fraundorfer, ``Visual odometry [tutorial],'' \emph{IEEE robotics \& automation magazine}, vol.~18, no.~4, pp. 80--92, 2011.

\bibitem{ahn2012board}
S.~Ahn, S.~Yoon, S.~Hyung, N.~Kwak, and K.~S. Roh, ``On-board odometry estimation for 3d vision-based slam of humanoid robot,'' in \emph{2012 IEEE/RSJ International Conference on Intelligent Robots and Systems}.\hskip 1em plus 0.5em minus 0.4em\relax IEEE, 2012, pp. 4006--4012.

\bibitem{wirbel2013humanoid}
E.~Wirbel, B.~Steux, S.~Bonnabel, and A.~de~La~Fortelle, ``Humanoid robot navigation: From a visual slam to a visual compass,'' in \emph{2013 10th IEEE international conference on networking, sensing and control (ICNSC)}.\hskip 1em plus 0.5em minus 0.4em\relax IEEE, 2013, pp. 678--683.

\bibitem{gouda2013vision}
W.~Gouda, W.~Gomaa, and T.~Ogawa, ``Vision based slam for humanoid robots: A survey,'' in \emph{2013 Second International Japan-Egypt Conference on Electronics, Communications and Computers (JEC-ECC)}.\hskip 1em plus 0.5em minus 0.4em\relax IEEE, 2013, pp. 170--175.

\bibitem{scona2017direct}
R.~Scona, S.~Nobili, Y.~R. Petillot, and M.~Fallon, ``Direct visual slam fusing proprioception for a humanoid robot,'' in \emph{2017 IEEE/RSJ International Conference on Intelligent Robots and Systems (IROS)}.

\bibitem{stasse2006real}
O.~Stasse, A.~J. Davison, R.~Sellaouti, and K.~Yokoi, ``Real-time 3d slam for humanoid robot considering pattern generator information,'' in \emph{2006 IEEE/RSJ International Conference on Intelligent Robots and Systems}.

\bibitem{hourdakis2021roboslam}
E.~Hourdakis, S.~Piperakis, and P.~Trahanias, ``roboslam: Dense rgb-d slam for humanoid robots,'' in \emph{2021 IEEE/RSJ International Conference on Intelligent Robots and Systems (IROS)}.\hskip 1em plus 0.5em minus 0.4em\relax IEEE, 2021, pp. 2224--2231.

\bibitem{zhang2018dense}
T.~Zhang, E.~Uchiyama, and Y.~Nakamura, ``Dense rgb-d slam for humanoid robots in the dynamic humans environment,'' in \emph{2018 IEEE-RAS 18th International Conference on Humanoid Robots (Humanoids)}.

\bibitem{whelan2015elasticfusion}
T.~Whelan, S.~Leutenegger, R.~Salas-Moreno, B.~Glocker, and A.~Davison, ``Elasticfusion: Dense slam without a pose graph.''\hskip 1em plus 0.5em minus 0.4em\relax Robotics: Science and Systems, 2015.

\bibitem{tanguy2019closed}
A.~Tanguy, D.~De~Simone, A.~I. Comport, G.~Oriolo, and A.~Kheddar, ``Closed-loop mpc with dense visual slam-stability through reactive stepping,'' in \emph{2019 International Conference on Robotics and Automation (ICRA)}.\hskip 1em plus 0.5em minus 0.4em\relax IEEE, 2019, pp. 1397--1403.

\bibitem{sumikura2019openvslam}
S.~Sumikura, M.~Shibuya, and K.~Sakurada, ``Openvslam: A versatile visual slam framework,'' in \emph{Proceedings of the 27th ACM International Conference on Multimedia}, 2019, pp. 2292--2295.

\bibitem{rublee2011orb}
E.~Rublee, V.~Rabaud, K.~Konolige, and G.~Bradski, ``Orb: An efficient alternative to sift or surf,'' in \emph{2011 International conference on computer vision}.\hskip 1em plus 0.5em minus 0.4em\relax Ieee, 2011, pp. 2564--2571.

\bibitem{kummerle2011g}
R.~K{\"u}mmerle, G.~Grisetti, H.~Strasdat, K.~Konolige, and W.~Burgard, ``g 2 o: A general framework for graph optimization,'' in \emph{2011 IEEE International Conference on Robotics and Automation}.\hskip 1em plus 0.5em minus 0.4em\relax IEEE, 2011.

\bibitem{9663522}
A.~Vedadi, K.~Sinaei, P.~Abdolahnezhad, S.~S. Aboumasoudi, and A.~Yousefi-Koma, ``Bipedal locomotion optimization by exploitation of the full dynamics in dcm trajectory planning,'' in \emph{2021 9th RSI International Conference on Robotics and Mechatronics (ICRoM)}, 2021.

\bibitem{10000072}
P.~Abdolahnezhad, A.~Yousefi-Koma, A.~Vedadi, K.~Sinaei, B.~Maleki, and M.~Shafiee, ``Online bipedal locomotion adaptation for stepping on obstacles using a novel foot sensor,'' in \emph{2022 IEEE-RAS 21st International Conference on Humanoid Robots (Humanoids)}, 2022, pp. 344--349.

\end{thebibliography}

\end{document}